\documentclass{article}
\usepackage{amsmath,graphicx,mlspconf}
\usepackage{xcolor}
\usepackage{arabtex}
\usepackage{multirow}
\usepackage{arydshln}
\usepackage{babel}
\usepackage{placeins}
\usepackage{pifont}
\usepackage{hyperref}
\usepackage{natbib}

\copyrightnotice{979-8-3503-2411-2/25/\$31.00 {\copyright}2025 IEEE}

\title{SAGE: Spliced-Audio Generated Data for Enhancing Foundational Models in Low-Resource Arabic-English Code-Switched Speech Recognition}
\name{Muhammad Umar Farooq, Oscar Saz} %\thanks{Anonymous.}
\address{Emotech Ltd., UK}

\begin{document}
\ninept

\maketitle

\begin{abstract}
This paper investigates the performance of various speech SSL models on dialectal Arabic (DA) and Arabic-English code-switched (CS) speech. To address data scarcity, a modified audio-splicing approach is introduced to generate artificial CS speech data. Fine-tuning an already fine-tuned SSL model with the proposed Spliced-Audio Generated (SAGE) data results in an absolute improvement on Word Error Rate (WER) of 7.8\% on Arabic and English CS benchmarks. Additionally, an Experience Replay (ER) inspired approach is proposed to enhance generalisation across DA and CS speech while mitigating catastrophic forgetting. Integrating an out-of-domain 3-gram language model reduces the overall mean WER from 31.7\% to 26.6\%. Few-shot fine-tuning for code-switching benchmarks further improves WER by 4.9\%. A WER of 31.1\% on Arabic-English CS benchmarks surpasses large-scale multilingual models, including USM and Whisper-large-v2 (both over ten times larger) by an absolute margin of 5.5\% and 8.4\%, respectively.
\end{abstract}
\begin{keywords}
speech recognition, code-switching, low-resource, foundation models, dialectal Arabic

\end{keywords}

\section{Introduction}
\label{sec:intro}
%Code-switching (CS), the alternation between two or more languages within a single conversation, is prevalent in multicultural, multilingual societies and regions.
%The primary language in such exchanges is known as the matrix language, while the secondary language is called the embedded language.
%This phenomenon poses significant challenges for monolingual ASR systems when encountering embedded languages.

%Code-switching (CS), the alternation between two or more languages within a single conversation, is common in multicultural, multilingual societies. This phenomenon presents significant challenges for monolingual ASR systems when encountering embedded languages.
Code-switched speech, where speakers mix one or more embedded languages into their first language, poses significant challenges for traditional monolingual Automatic Speech Recognition (ASR) systems due to their inability to deal effectively with words or sentences from an unseen language.
Bilingual ASR systems improve performance \cite{nguyen2022} but still fall short compared to systems trained on actual code-switched data \cite{mustafa22}.
%Bilingual ASR systems improve this performance , but they cannot match systems with trained with actual CS data \cite{mustafa22}.
While modern large multilingual models such as Whisper-large \cite{whisper}, Universal Speech Model (USM) \cite{googleusm} or Massively Multilingual Speech (MMS) \cite{metamms} may seem like ideal solutions, their performance remains strong for resource-rich languages but degrades for limited data resources.

Although code-switching has become more widespread with globalisation \cite{ye22}, code-switched speech data remains scarce \cite{du2021} due to the challenges involved in its collection. While code-switching has received more attention over the last few years \cite{farooq20, ogunremi23, ugan24}, research has been limited to only a few language pairs \cite{seame,chan05b,mucs}.
%such as English-Mandarin\cite{seame}, Cantonese-English \cite{chan05b} and sesveral Indian languages \cite{mucs}.
In previous work, various data augmentation techniques such as speech editing \cite{du2021}, audio splicing and neural TTS \cite{liang23b} have been used to overcome data scarcity of code-switched speech. Different splicing approaches have been explored for code-switched data augmentation, such as acoustic feature splicing \cite{ye22} and concatenating small speech units \cite{liang23b}. Though the previous work has shown the gains by augmenting spliced audio data, it has only been limited to Mandarin-English code-switched speech \cite{ye22, du2021, liang23b}.% and not very well explored yet. 

With the recent advances in the domain of Self-Supervised Learning (SSL) models, speech foundation models have been explored to evaluate their capabilities in many different ASR tasks such as cross-lingual \cite{w2v-xlsr}, cross-domain \cite{ahmad24}, low-resource \cite{zhu22c}, noisy background \cite{song23} and streaming speech recognition \cite{fu24}. 
However, limited work has been done to assess the capabilities of different SSL models for code-switched speech recognition \cite{huang24}. Code-switched speech presents unique challenges for SSL models, and their performance can vary depending on the target and pretraining languages.
%However, very limited work has been done on gauging the capabilities of different SSL models for code-switched speech recognition \cite{huang24} which poses its challenges for SSL models and can depend on target languages and pretraining languages.

Arabic, a diglossic language \cite{zaebuc}, has a formal Modern Standard Arabic (MSA) and many regional dialects such as Egyptian, Saudi, Gulf, Moroccan, etc. Moreover, French and English are commonly mixed with Arabic by native speakers (of North Africa and the Middle East, respectively) in daily conversations due to historical background \cite{arzen}. So, inter-dialectal and Arabic-English code-switching is a common phenomenon in many Middle Eastern countries. However, most of the work done for Arabic speech technologies or resources, e.g. CommonVoice (CV) \cite{cv}, MGB-2 \cite{mgb2} and QASR \cite{qasr} has primarily focused on MSA dialect, leaving a research gap for unbiased and fair systems for dialectal Arabic. Recently, new small-scale resources for Arabic-English code-switched speech, ZAEBUC \cite{zaebuc} and ArzEn \cite{arzen}, have been developed. However, the research in speech technologies for CS in Arabic dialects is still sparse.
%Furthermore, building an ASR system for dialectal Arabic which could also recognise Arabic-English code-switched speech is very challenging. Since the model is supposed to learn many different tasks, smaller models can suffer from learning all the tasks fairly.

%Nowadays, massive amounts of training data are used for training of large multilingual models such as Whisper \cite{whisper}, Universal Speech Model (USM) \cite{googleusm} or Massively Multilingual Speech (MMS) \cite{metamms} models. These models usually perform well on rich resource languages but the performance degrades for the languages on the tail end in terms of data resources. Most of the Arabic data resources such as CommonVoice (CV) \cite{cv}, MGB-2 \cite{mgb2} and QASR \cite{qasr} mainly consist of the MSA dialect. For Arabic-English code-switched speech, recent ZAEBUC \cite{zaebuc} and ArzEn \cite{arzen} data sets are around 10-12 hours which is not enough to train a reasonably well performing ASR system.
%Additionally, fine-tuning some large SSL models such as Whisper-large (or even small and medium) and XLSR is computationally expensive as well.

Given the aforementioned research gaps for DA, inter-dialectal and Arabic-English CS speech, the key contributions of this work include:
\begin{itemize}
    \item Evaluating various base-scale foundational models (of up to 100M parameters) to train a fair and unbiased ASR system for multi-dialectal Arabic with Arabic-English code-switching.
    \item Employing the ER \cite{rolnick19} inspired approach for continual learning of a smaller model to avoid catastrophic forgetting.
    \item Proposing a modified audio splicing approach to generate Arabic-English code-switched data and using SAGE data for improving SSL models' performance
\end{itemize}

Incorporating the proposed SAGE data improves mean WER by 7.8\% on Arabic-English CS benchmarks, while an ER-inspired fine-tuning approach enhances generalisation across DA and CS benchmarks. Integrating a 3-gram language model further lowers the overall mean \%WER by 5.1\%. For code-switching datasets of ZAEBUC and ArzEn dev and test sets, a few-shot fine-tuning further reduces WER from 36.0\% to 31.1\%, outperforming larger multilingual models like USM, MMS, and Whisper-large-v2. These findings underscore the challenges of code-switched speech for large multilingual models.

\section{Spliced-Audio GEnerated (SAGE) Data}
\label{sec:sage}

\begin{figure}[t]
  \centering
  \includegraphics[width=\linewidth]{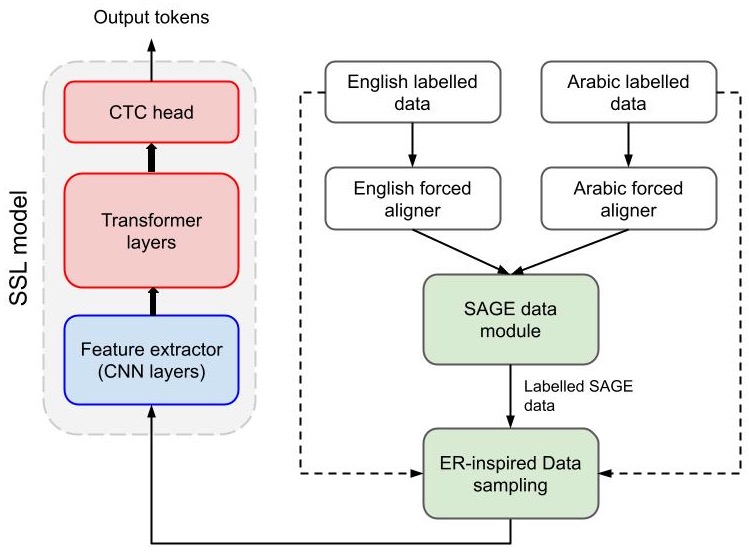}
  \caption{Fine-tuning of SSL models using SAGE data, with only the red-outlined blocks being fine-tuned.}
  \label{fig:flow}
\end{figure}

%Due to the scarcity of code-switched speech data, various data augmentation techniques such as speech editing and audio splicing have been explored \cite{liang23b, du2021}.
In previous work like \cite{liang23b}, Mandarin-English code-switched speech data has been generated by splicing audio of the primary language and TTS-generated audio of the secondary language. However, TTS-generated speech still presents acoustic differences from genuine speech, hampering the learning of ASR systems \cite{su2024}. The split-and-splice approach \cite{du2021} splits natural code-switched utterances into language-specific segments, with new data generated by splicing random segments from the same speaker. However, since this data has been solely generated from code-switched data, its quantity remains constrained by the size of the original dataset. A previous approach \cite{hussein2024speech} to generating Arabic-English code-switched data involved creating audio by splicing word segments from different recordings, guided by a code-switched text. This text has been generated from a parallel Arabic-English corpus. However, this method is relatively costly, as it depends on the creation of a parallel text corpus, which in turn requires a machine translation system.

In this work, it is proposed to generate Arabic-English code-switched speech data by splicing segments from monolingual sources. Word boundaries for Arabic and English are estimated using monolingual ASR systems and Viterbi alignment. To generate a code-switched sample, base and fragment audio segments are randomly selected from different languages, i.e. Arabic and English or vice versa. Using alignment information, a fragment containing two to four words is randomly extracted from the fragment audio and inserted at a selected insertion point within the base audio segment. The insertion point is chosen randomly using base audio alignment information and can occur between two words, at the start, or at the end of the sentence. The resulting spliced audio segment is volume-normalised, and the insertion points are smoothed for seamless transitions. This spliced-audio generated data (also referred to as SAGE data) is used to fine-tune a speech foundation model with an ER-inspired data sampling approach as shown in Figure \ref{fig:flow}. SAGE data has multiple advantages over previous work;

\begin{itemize}
    %\item By generating code-switched samples in both ways, English within Arabic and Arabic within English, SAGE data generalises better and maintains optimal performance for both languages.
    \item By generating code-switched samples both ways—English within Arabic and vice versa—SAGE data \textbf{generalises better} and maintains optimal performance for both languages.
    %Naturally, speakers code-switch in both ways and the results in Section \ref{ss:res.sage} have shown that the model is better generalised for either type of code-switching when fine-tuned on SAGE data.
    \item Since \textbf{monolingual resources} of both languages are utilised, it allows for the generation of unrestricted amounts of data.
    \item Only \textbf{natural speech segments} are used instead of TTS voices, mitigating the downsides of the synthetic data.
\end{itemize}

SAGE data is generated by splicing audio segments without ensuring they come from the same speaker, which could result in speaker mismatches within an utterance. This may appear detrimental to the training of a model. Nevertheless, the focus of this work is to utilise the SAGE data to fine-tune an SSL model for the downstream task of speech recognition. The previous work shows that the top layers of SSL encoders generate speaker-independent embeddings and focus more on semantic information \cite{pasad23}, thus mitigating the risk of splicing multiple speakers in a single utterance. So, an SSL model is expected to learn code-switching from SAGE data without focusing on speaker information.

\begin{figure}[t]
  \centering
  \includegraphics[width=\linewidth]{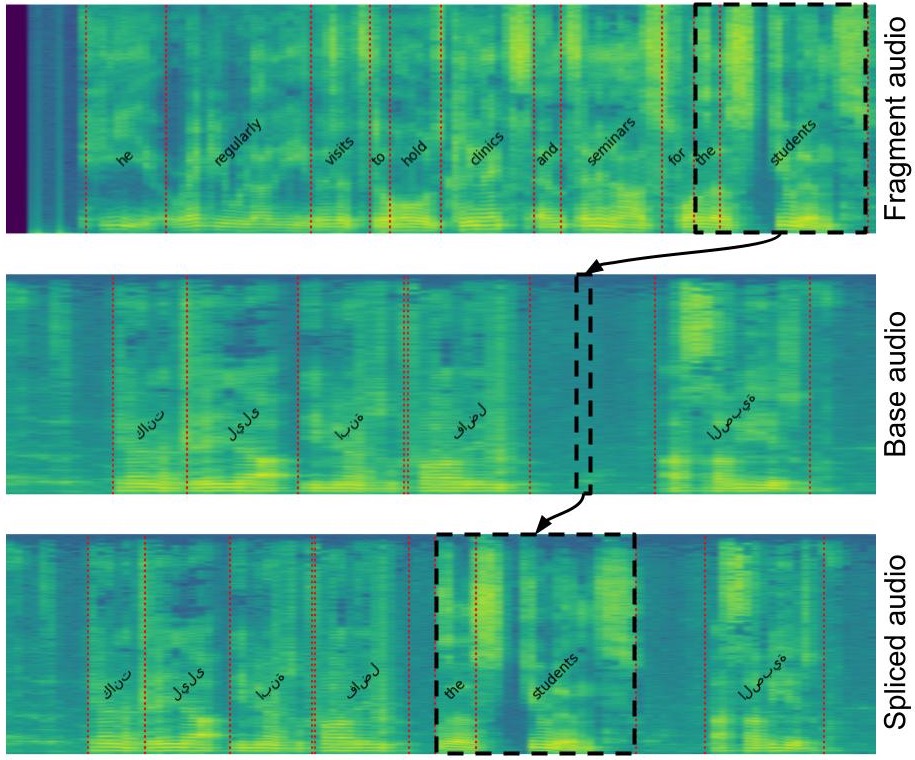}
  \caption{Spectrograms of fragment, base and spliced (SAGE) audio samples with alignments of a random sample.}
  \label{fig:sample}
\end{figure}

In Figure \ref{fig:sample}, spectrograms of the base audio, fragment audio and the resulting spliced audio of an example are shown. The fragment ``the students" from the fragment audio of an English sentence ``He regularly visits to hold clinics and seminars for the students" is inserted in the base audio with the Arabic sentence ``\<كانت ليلى ابنة فاضل الصبية>", resulting in the completely new sentence ``\<الصبية> the students \<كانت ليلى ابنة فاضل>".
%The spectrograms contain word alignment and they show where the English fragment ``the students" has been inserted between the penultimate and last words of the base Arabic sentence.

% \begin{figure}[t]
%   \centering
%   \includegraphics[width=1.00\linewidth]{codeswitch_ex.png}
%   \caption{Artificial code-switch sample with alignment}
%   \label{fig:sample}
% \end{figure}

%\section{Dialectal Arabic and English code-switching / SSL Models and CS / Details on Franken data}
% \begin{table*}[!htbp]
%     \centering
%     \caption{Caption}
%     \begin{tabular}{l|c|ccc|cccc}
%         \hline
%         \multirow{3}{*}{Model}&\multirow{3}{*}{\# param (M)}&\multicolumn{3}{c|}{Dialectal Arabic}&\multicolumn{4}{c}{Ar-En CS}\\
%         \cline{3-5}
%         \cline{6-9}
%         && \multirow{2}{*}{CV} & \multirow{2}{*}{FLEURS} & \multirow{2}{*}{MGB2} & \multicolumn{2}{c}{ZAEBUC} & \multicolumn{2}{c}{ArzEn} \\
%         \cline{6-9}
%         &&&&&dev&test&dev&test\\
%         \hline\hline
%         w2v-base&97.3&17.1&22.7&32.7&44.0&43.8&56.8&57.2\\
%         Hubert-base&97.3&17.4&22.7&31.5&51.8&48.4&53.0&53.3\\
%         mHubert&97.3&14.4&19.3&26.6&47.7&45.4&50.1&50.4\\
%         Whisper-base&74&&&&&&&\\
%         % Whisper-small&244&&&&&&&\\
%         w2v-XLSR&318.7&15.5&23.2&35.0&64.7&58.9&53.4&55.6\\
%         \hline
%     \end{tabular}
%     \label{tab:res.ssl}
% \end{table*}

\section{Data}
\label{sec:data}

\begin{table*}[t]
    \centering
    \caption{WER(\%) for foundational models on dialectal Arabic and code-switched after fine-tuning on 1217.26 hours of Arabic and English data}
    \begin{tabular}{l|c||ccc|c||cccc|c||c}
        \hline
        % \multirow{2}{*}{Model}&\multirow{2}{*}{\# param (M)}&\multicolumn{3}{c}{Dialectal Arabic}\\
        % \cline{3-5}
        % && CV & Fleurs & MGB2 \\
        \multirow{2}{*}{Model}&\multirow{2}{*}{\#params}&\multirow{2}{*}{CV} &\multirow{2}{*}{Fleurs}&\multirow{2}{*}{MGB2}&\multirow{2}{*}{DA avg.}& \multicolumn{2}{c}{ZAEBUC} & \multicolumn{2}{c|}{ArzEn} & \multirow{2}{*}{CS avg.}&\multirow{2}{*}{Overall avg.} \\
        &&&&&&dev&test&dev&test&&\\
        \hline\hline
         Whisper-base&75.3&30.7&35.8&47.4&40.4&66.1&63.5&66.7&68.3&66.3&52.2\\
        wav2vec2.0-base&97.3&17.1&22.7&32.7&26.2&\textbf{44.0}&\textbf{43.8}&56.8&57.2&50.3&37.2\\
        HuBERT-base&97.3&17.4&22.7&31.5&25.6&52.8&48.7&53.0&52.2&51.9&37.6\\
        mHuBERT-147&97.3&\textbf{14.4}&\textbf{19.3}&\textbf{26.6}&\textbf{21.6}&48.2&45.8&\textbf{50.1}&\textbf{50.5}&\textbf{48.8}&\textbf{33.9}\\
        XLSR&318.7&11.9&21.1&31.0&23.2&64.5&59.1&53.0&54.4&58.3&39.1\\
        \hline
    \end{tabular}
    \label{tab:res.ssl_da}
\end{table*}
The experiments for the development and evaluation of the proposed approaches use multiple sources of Arabic and English data. The monolingual English data, used for training, is 449.2 hours from the following sources:

\begin{itemize}
    \item AMI corpus \cite{ami2007}: A collection of 93 hours of in-person meeting recordings in different locations. Only the head-mounted microphone speech is used.
    \item English CommonVoice \cite{commonvoice2020}: A subset of 136.5 hours of English speech from the CV English train set.
    \item TED-LIUM2 \cite{tedlium2012}: A collection of 219.7 hours of TED talks collected and transcribed from the TED website.
\end{itemize}

For Arabic, 768.1 hours of multi-dialect Arabic from the following sources are used:

\begin{itemize}
    \item Arabic CommonVoice \cite{commonvoice2020}: A subset of 102 hours of Arabic speech recorded as part of the CommonVoice project.
    \item Conversational dialectal speech\footnote{Due to the limited availability of public Arabic dialectal speech data, this work relies heavily on proprietary sources. The authors are open to collaboration with research institutions interested in using this data.}: A total of 666.1 hours of conversations featuring Saudi, Egyptian, and Gulf speakers in their respective colloquial dialects were collected through telephony, mobile, and web applications. This dataset includes 398.2 hours in the Saudi dialect, 249.2 hours in the Egyptian dialect, and 18.7 hours in the Gulf dialect.
\end{itemize}

Finally, the two code-switching speech corpora that are used for evaluation are described in more detail next.

\subsection{ZAEBUC-spoken corpus}
\label{sec:data.zaebuc}

The ZAEBUC-spoken speech corpus \cite{zaebuc} (commonly referred to as ZAEBUC in this paper) is a collection of multilingual, multidialectal web meetings where multiple Arabic participants discuss their brainstormed ideas with an interlocutor in four phases. These phases ensure English-only, Arabic-only and Arabic-English code-switched utterances. Both Arabic and English participants are from different countries, speaking different dialects of both languages. For experimentation, the ZAEBUC dataset is split into 4.66 hours of development set and 2.83 hours of evaluation set, with a small amount of 2 hours used for few-shot learning experiments. 

\subsection{ArzEn corpus}
\label{sec:data.arzen}

The ArzEn speech corpus \cite{arzen} is a collection of 12 hours of Egyptian Arabic-English code-switched corpus recorded by 38 bilingual students. The recordings are informal 1-on-1 interviews in a soundproof room. For experimental purposes, the data set has been split into train, development and test sets of 5.6, 2.9 and 2.9 hours, respectively \cite{arzen}. The train set is used for few-shot learning experimentation, while the development and test sets are used for evaluation purposes.

\subsection{Data preparation}

All sources of data are pre-processed in the same way before experimentation. All audio files are converted to 16-bit WAV files with a sampling rate of 16,000 Hz.
%large conversational files are split into individual segment audio files based on the timestamps provided by each corpus.
All sources of text are normalised according to standardised rules for English and Arabic, with numbers, symbols and acronyms converted to text form and all punctuation stripped. The English text is converted to the basic 26-character lowercase set plus the apostrophe. Arabic text is stripped of all diacritics and characters are converted to their isolated Unicode form. The final Arabic character set used is the basic 28 Arabic letters plus the 2 modified letters (\textit{ta marbuta} and a\textit{lif maqsura}) and 6 \textit{Hamza} forms (\textit{hamza}, \textit{alif hamza} above, \textit{alif hamza} below, \textit{alif madda}, \textit{waw hamza} and \textit{ya hamza}).

\subsection{Language Modelling}
\label{ss:es.lm}
Most of the results in this work are reported without the use of a language model unless explicitly specified. For the results with an LM, a 3-gram language model is integrated, which is trained on a text corpus from the fine-tuning corpus of only monolingual sources, of both Arabic and English, described earlier. The KenLM toolkit \cite{kenlm} is used to build the language model.

\section{Evaluation of foundational models}
\label{sec:setup}

Initial experiments are performed to evaluate the dialectal Arabic and Arabic-English code-switched speech recognition capabilities of different foundational models. For this purpose, all monolingual datasets, mentioned in Section \ref{sec:data} (both Arabic and English), are pooled together for a total of 1,217.26 hours of training data. A total of 12.9 hours of speech is used for validation, consisting of 8.2 hours of held-out data from the conversational dialectal Arabic datasets and the 4.7 hours of development set of ZAEBUC. Evaluation is conducted on the test sets of three common Arabic benchmarks, i.e. Arabic CommonVoice, FLEURS, and MGB2, and the development and test sets of both ZAEBUC and ArzEn as CS benchmarks. Although Arabic CV data is mostly MSA dialect, MGB-2 contains different Arabic dialects, and FLEURS is also MSA but spoken by native Egyptian speakers.

To evaluate the performance of foundational models on this setup, several foundational models are experimented with. In addition to their performance on downstream tasks, model size is also crucial, as larger models require more training time, memory, and computational resources. For that, the focus here is on comparing foundational models of base-scale (less than 100 million parameters). With that in mind, various base-scale speech foundational models, including HuBERT base \cite{hubert}, wav2vec2.0 base \cite{w2v2}, and Whisper base \cite{whisper}, are compared. A multilingual version of HuBERT pretrained on multilingual data (mHuBERT-147) \cite{mhubert} is also evaluated to understand the importance of pretraining data. And, finally, to compare against a larger model, the XLSR pretrained model \cite{w2v-xlsr} with over 300M parameters is also evaluated. 

\begin{table*}[t]
    \centering
    \caption{WER(\%) of ER fine-tuned mHubert model on dialectal data. $\mathcal{M}_{X}^{Y}$ is a model $\mathcal{M}$ fine-tuned on total $X$ hours from $Y$ data sets.}
    \begin{tabular}{l||ccc|c||cccc|c||c}
        \hline
        \multirow{2}{*}{Model}&\multirow{2}{*}{CV}&\multirow{2}{*}{Fleurs}&\multirow{2}{*}{MGB2}&\multirow{2}{*}{DA avg.}& \multicolumn{2}{c}{ZAEBUC} & \multicolumn{2}{c|}{ArzEn}&\multirow{2}{*}{CS avg.}&\multirow{2}{*}{Overall avg.}\\
        &&&&&dev&test&dev&test&&\\
        \hline\hline
        Bashar et al. \cite{talafha23_interspeech}&22.2&23.7&29.7&25.2&-&-&-&-&-&-\\
        \hdashline
        $\mathcal{M}$&14.4&19.3&26.6&21.6&48.2&45.8&50.1&50.5&48.8&33.9\\        $\mathcal{M}_{250}^{SAGE}$&19.9&32.3&21.6&26.9&35.2&35.2&47.0&47.3&41.0&33.3\\
        $\mathcal{M}_{450}^{Ar, En, SAGE}$&14.6&29.0&18.7&22.9&37.3&37.3&47.6&47.2&42.1&31.7\\
        $\mathcal{M}_{300}^{Ar, En, SAGE}$&14.1&29.1&18.8&22.8&37.7&37.9&47.5&47.1&42.4&31.7\\
        ~ + LM&\textbf{11.7}&\textbf{22.0}&\textbf{16.0}&\textbf{17.8}&\textbf{31.4}&\textbf{31.7}&\textbf{43.0}&\textbf{43.4}&\textbf{37.1}&\textbf{26.6}\\
        % \fofo{E4}&15.9&20.5&31.0&22.5&34.7&35.1&46.5&46.4&40.7\\
        \hline
    \end{tabular}
    \label{tab:res.franken_da}
\end{table*}

All the foundational models are fine-tuned on the aforementioned 1,217.26 hours of Arabic and English speech for the downstream task of speech recognition. For fine-tuning, a model head of four linear layers is applied on top of the encoder of the SSL model. The dimensionality of the output of the last layer is the same as the number of unique characters in the dataset, along with a space and a blank character. The CTC objective is used to fine-tune the whole model. All models are fine-tuned for 40 epochs, and the best epoch in terms of WER on the dev set is used for evaluation. All the experiments are done using the SpeechBrain \cite{speechbrain} toolkit.

\begin{table*}[]
    \centering
    \caption{WER(\%) of Arabic-only, English-only and code-switched utterances of code-switched data sets (ZAEBUC and Arzen dev and test sets) -\textit{ WER is shown as (Ar / En / CS)}}
    \begin{tabular}{l|cccc}
        \hline
        \multirow{2}{*}{Exp} & \multicolumn{2}{c}{ZAEBUC} & \multicolumn{2}{c}{ArzEn} \\
        \cline{2-3}
        \cline{4-5}
        &dev&test&dev&test\\
        \hline \hline
        $\mathcal{M}$ & 36.1 / 50.1 / 54.0 & 38.1 / 43.1 / 61.7 & 43.0 / 30.8 / 52.3 & 44.4 / 100.0 / 52.4 \\
        $\mathcal{M}_{250}^{SAGE}$ & 36.4 / \textbf{28.4} / \textbf{45.7} & 37.4 / \textbf{29.0} / \textbf{46.1} & 45.5 / \textbf{30.8} / \textbf{47.5} & 46.6 / 100.0 / \textbf{47.5} \\
        $\mathcal{M}_{450}^{Ar, En, SAGE}$ & \textbf{34.4} / 32.3 / 47.8 & \textbf{36.0} / 32.1 / 50.5 & 40.4 / 46.2 / 49.8 & 42.4 / 106.1 / 48.6\\
        $\mathcal{M}_{300}^{Ar,En,SAGE}$ & 34.7 / 32.7 / 48.4 & 36.1 / 33.0 / 51.3 & \textbf{39.8} / 53.8 / 49.8 & \textbf{42.3} / 100.0 / 48.6\\
        \hdashline
        ~ + LM & 26.7 / 27.7 / 41.2 & 28.6 / 27.7 / 44.5 & 36.6 / 46.2 / 45.0 & 39.4 / 69.7 / 44.6\\
        \hline
    \end{tabular}
    \label{tab:res.utt}
\end{table*}

Table \ref{tab:res.ssl_da} presents the results of the different foundational models for both dialectal Arabic and code-switched Arabic-English evaluation sets. For DA benchmarks, the mHuBERT-147 model outperforms all other foundational models, achieving a mean WER of 21.6\%. In the case of Arabic-English code-switched data, wav2vec2.0 provides the best results for ZAEBUC dev and test sets, but mHuBERT-147 is best for ArzEn dev and test sets. This difference in performance is found to be due to the larger amount of English words in the ZAEBUC dataset, where over 60\% of the speech is English, compared to ArzEn, which only contains 15\% of English words. However, on average, mHubERT-147 is the best performer with 48.8\% WER on CS benchmarks and 33.9\% as an overall WER on all the datasets. With these results, the model fine-tuned on mHuBERT-147 is selected for further experiments. % for code-switching ASR.

\section{Results and Discussion}
\label{sec:results}

%\subsection{SSL Models}
%\label{ss:rd.ssl}

%As described in Section \ref{ss:es.ssl}, various SSL models are fine-tuned on Arabic and English data sets. Although the primary focus is on base-scale SSL models (with a number of parameters less than 100 million), the wav2vec2.0 based XLSR model is also experimented with which is a small-scale SSL model (with around 317 million parameters). In the experimentation here, base-scale SSL models include Hubert-base, wav2vec2.0-base, Whisper-base and mHubert whereas XLSR is a small-scale model. Out of these models, only wav2vec2.0-base (w2v-base) and Hubert-base are monolingual models and the rest of them are pre-trained on multilingual data sets. All of these models are fine-tuned on 1217.26 hours of Arabic and English data sets described in \ref{sec:data}. All fine-tuned models are evaluated on both, dialectal Arabic and Arabic-English code-switched corpora. The results for DA and code-switched benchmarks are shown in Table \ref{tab:res.ssl_da} and Table \ref{tab:res.ssl_cs} respectively.

% \subsection{SAGE data}
% \label{ss:res.sage}

%Further experimentation is done to improve the performance of the best SSL model (mHubert) for Arabic inter-dialectal and Arabic-English code-switching.
%As described in Section \ref{sec:sage}, SA data is generated and used to further enhance the model capabilities to recognise code-switched speech.

SAGE data (as described in Section \ref{sec:sage}) is used to further enhance the model's capabilities to recognise code-switched speech. The objective here, however, is to enhance the model's ability to recognise code-switched utterances while retaining its knowledge of dialectal Arabic speech with minimal distortion. Initially, 250 hours of SAGE data is used for further fine-tuning of the model. The results are shown in Table \ref{tab:res.franken_da} for dialectal Arabic and Arabic-English code-switched speech benchmarks. The first row shows the performance of the Whisper-small model from the previous work \cite{talafha23_interspeech} on DA benchmarks. The baseline results of mHubert-147 (from Table \ref{tab:res.ssl_da}) are shown in the first row after the dashed line  ($\mathcal{M}$). The performance of the model, further fine-tuned on 250 hours of SAGE data, is shown as $\mathcal{M}_{250}^{SAGE}$.
%It is worth mentioning that no in-domain data from evaluation benchmarks (ZAEBUC and ArzEn) is used in training so far. So, the results are considered as zero-shot performance of the model. 

It can be seen that fine-tuning the $\mathcal{M}$ model (mHubert-147 from Table \ref{tab:res.ssl_da}) on SAGE data significantly reduces the WER on code-switched data sets. Though it yields a mean absolute reduction of 7.8\% across CS benchmarks, it increases the WER for dialectal Arabic evaluation test sets by a mean absolute of 5.3\%. To overcome this shortcoming of catastrophic forgetting, experiments are done with an ER-inspired approach. So, the model is fine-tuned on SAGE data mixed with the already-learned Arabic-only and English-only data sets. For $\mathcal{M}_{450}^{Ar, En, SAGE}$, equal amounts of 100 hours from Arabic and English are mixed with 250 hours of SAGE data to fine-tune the $\mathcal{M}$ model. This reduces the DA error to 22.9\% with a minor degradation of 1.1\% in mean WER for code-switched datasets. As a study to determine whether the performance gains stem from SAGE data or simply from the larger volume of data, the 200 hours of Arabic-only and English-only data are fixed, and the SAGE data is also reduced to 100 hours. Fine-tuning $\mathcal{M}$ on this mix 300 hours ($\mathcal{M}_{300}^{Ar,En,SAGE}$) causes less degradation for DA data sets but marginally increases CS error to 42.4\%. Though this 300 hours of data is more than 250 hours of SAGE data ($\mathcal{M}_{250}^{SAGE}$), $\mathcal{M}_{250}^{SAGE}$ reduces the CS error rate by 1.4\% absolute when compared with $\mathcal{M}_{300}^{Ar,En,SAGE}$. It implies that the SAGE data helps the model perform well on CS data but makes it prone to catastrophic forgetting. So, fine-tuning the SSL model using an experience replay approach results in a more generalised model ($\mathcal{M}_{300}^{Ar,En,SAGE}$).

To further analyse the behaviour of the SAGE data in model fine-tuning, the CS results are dissected into WER on Arabic only, English only and code-switched utterances. From Table \ref{tab:res.utt}, it can be seen that fine-tuning on only synthetic data ($\mathcal{M}_{250}^{SAGE}$) decreases the WER massively for English-only and CS utterances, e.g. an absolute reduction of 14.1\% and 15.6\% is yielded for English-only and CS utterances of the ZAEBUC-test set. However, the error rate for Arabic-only utterances is slightly increased across almost all the data sets. The performance of Arabic-only utterances can also be improved by either mixing the Arabic-only and English-only utterances in the fine-tuning stage ($\mathcal{M}_{450}^{Ar,En,SAGE}$) and reducing the amount of synthetic data in the mixed data ($\mathcal{M}_{300}^{Ar,En,SAGE}$). However, reducing the synthetic data reduces performance on English-only and CS utterances. The data set statistics show that the ArzEn dev and test sets have very few English-only utterances (only 33 and 13 reference words, respectively). So, due to very short utterances and very few data samples, the WER for the ArzEn test set is 100\% or more than that in Table \ref{tab:res.utt}. For the analysis, the error rate of the Arzen data sets is ignored. The last row shows the results when an OOD 3-gram language model is integrated, which reduces the WER for all data sets (ignoring \textit{En} results of ArzEn sets).

\begin{table}[b]
    \centering
    \caption{Performance of mHubert-147 model further fine-tuned on in-domain data of Arabic-English code-switched data sets. $\mathcal{FT}_{X}^{Y}$ is the model $\mathcal{M}^{SAGE}_{250}$ further finetuned on X hours from Y datasets.}
    \begin{tabular}{l|cccc|c}
        \hline
        \multirow{2}{*}{Model}&\multicolumn{2}{c}{ZAEBUC}&\multicolumn{2}{c|}{ArzEn}&\multirow{2}{*}{avg.}\\
        \cline{2-5}
        &dev&test&dev&test\\
        \hline
        MMS \cite{metamms}&49.8&45.9&83.7&84.5&65.6\\
         USM \cite{googleusm}&26.4&25.7&47.9&47.9&36.6\\
         Whisper large-v2 \cite{whisper}&25.9&27.1&53.5&53.9&39.5\\
         \hdashline
         $\mathcal{M}_{250}^{SAGE}$&28.0&28.0&47.4&41.6&36.0\\
         % ~ + LM &27.9 & & & \\
        $\mathcal{FT}_{1}^{ArzEn}$ &28.4&28.5&40.0&40.7&34.2\\ % + 1 hr Arzen
         $\mathcal{FT}_{1}^{ZAEBUC}$&26.3&26.7&44.3&44.8&35.2\\ % + 1 hr ZAEBUC
         $\mathcal{FT}_{2}^{ArzEn, ZAEBUC}$&\textbf{25.0}&\textbf{25.4}&\textbf{37.0}&\textbf{37.9}&\textbf{31.1}\\ % + 1 hr Arzen + ZAEBUC
         % ~ + LM &\textbf{25.2}&\textbf{25.4}&\textbf{37.0}&\textbf{37.9}&\textbf{31.2}\\
         $\mathcal{FT}_{Full}^{ArzEn, ZAEBUC}$&26.4&26.5&38.3&39.0&32.3\\

         \hline
    \end{tabular}
    \label{tab:res.fewshot}
\end{table}

Results from Table \ref{tab:res.franken_da} show that the model $\mathcal{M}_{300}^{Ar,En,SAGE}$ is well-generalised for both DA and Arabic-English code-switched data.
%However, the model from \fofo{E1} outperforms all other experiments for dialectal Arabic and Arabic-English code-switched data sets.
For experimentation so far, no training data from ZAEBUC or ArzEn data sets has been used for fine-tuning. So, the results for these data sets here are considered zero-shot. In the next step, the best model for code-switched data ($\mathcal{M}_{250}^{SAGE}$) is further fine-tuned in few-shot settings. In Table \ref{tab:res.fewshot}, results are shown when $\mathcal{M}_{250}^{SAGE}$ is further fine-tuned on 1 hour of training data from ZAEBUC and ArzEn train. When $\mathcal{M}_{250}^{SAGE}$ is further fine-tuned on 1 hour of ArzEn data set, a marginal improvement is seen for ArzEn data sets. Fine-tuning on 1 hour of ZAEBUC data set yields significant improvements for ZAEBUC datasets, but the performance is degraded for ArzEn datasets. An experiment is also done to fine-tune $\mathcal{M}_{250}^{SAGE}$, combining 1 hour of training data from each of the ZAEBUC and ArzEn train sets. This model outperforms across all the data sets, yielding an overall mean gain of 4.9\% compared with the baseline model of $\mathcal{M}_{250}^{SAGE}$. As an ablation study to determine if these gains are due to more fine-tuning data, an experiment is done where $\mathcal{M}_{250}^{SAGE}$ is finetuned on combining full amounts of both ZAEBUC and ArzEn training datasets. However, the results ($\mathcal{FT}_{Full}^{ArzEn, ZAEBUC}$) show that the model performance degrades for all the evaluation sets. All the results are shown after integrating the 3-gram language model.

%%\footnote{\href{https://huggingface.co/facebook/mms-1b-all}{https://huggingface.co/facebook/mms-1b-all}}
The performance of the final model on Arabic-English code-switched benchmarks is compared with very large multilingual speech models, including MMS (1B), USM (2B) and Whisper large-v2 (1.5B) models. The comparison of our best mHubert-147 model (97.3M), with the top three rows in Table \ref{tab:res.fewshot} shows that our fine-tuned model outperforms all the massive models across all the datasets. It also implies that the code-switched speech data is still a challenge faced by the massive models (either in terms of model capacity or covering more languages). Although USM and Whisper-large-v2 perform fairly well on ZAEBUC datasets, the performance of all the massive models degrades on ArzEn datasets, which are outperformed by even our zero-shot model ($\mathcal{M}_{250}^{SAGE}$ in Table \ref{tab:res.franken_da}).

% \begin{table*}[!htpb]
%     \centering
%     \caption{Caption}
%     \begin{tabular}{l|cccc}
%         \hline
%         \multirow{2}{*}{Exp} & \multicolumn{2}{c}{ZAEBUC} & \multicolumn{2}{c}{ArzEn} \\
%         \cline{2-3}
%         \cline{4-5}
%         &dev&test&dev&test\\
%         \hline \hline
%         \fofo{E0} & \textbf{36.1} / 50.1 / 54.0 & 38.1 / 43.1 / 61.7 & 43.0 / 30.8 / 52.3 & 44.0 / 100.0 / 52.4 \\
%         \fofo{E1} & 36.4 / \textbf{28.4} / \textbf{45.7} & 37.4 / \textbf{29.0} / \textbf{46.1} & 45.5 / \textbf{30.8} / \textbf{47.5} & 46.6 / 100.0 / \textbf{47.5} \\
%         \fofo{E2} & 34.3 / 32.7 / 48.2 & \textbf{36.0} / 32.1 / 50.5 & \textbf{41.0} / 46.2 / 49.5 & 42.8 / 103.0 / 48.6\\
%         \fofo{E3} & 34.7 / 32.7 / 48.4 & 36.1 / 33.0 / 51.3 & 39.8 / 53.8 / 49.8 & \textbf{42.3} / 100.0 / 48.6\\
%         \hline
%     \end{tabular}
%     \label{tab:res.utt}
% \end{table*}

\section{Conclusion}
\label{sec:conclusion}
This work evaluates various speech SSL models' capabilities for dialectal Arabic and Arabic-English code-switched speech. The mHubert-147 model (97.3M parameters) outperforms both similar-sized SSL models and the larger wav2vec2.0-based XLSR model (318.7M parameters). To address the lack of code-switched speech data, a modified audio-splicing approach (SAGE) is introduced, improving the model’s performance by 7.8\% on code-switched benchmarks. An experience replay-inspired data sampling approach enhances generalisation across CS and DA datasets. With the integration of an out-of-domain language model, the mean WER drops to 26.6\% from 31.7\%. A few-shot training on ZAEBUC and ArzEn data sets leads to a 5.9\% gain on CS benchmarks, surpassing large multilingual models like USM, MMS, and Whisper-large-v2. These results demonstrate that low-capacity models, when trained effectively, can be highly generalisable. The results also show that code-switched speech remains a challenge for large multilingual models, which cannot be overcome merely by expanding the number of languages or increasing model capacity.
% -------------------------------------------------------------------------

{\ninept
\bibliographystyle{unsrt}
\bibliography{refs}
}
\end{document}